# Optimizing Luxury Vehicle Dealership Networks: A Graph Neural Network Approach to Site Selection

Luca Silvano Carocci
*Nova School of Business and Economics*
*Universidade NOVA de Lisboa*
Carcavelos, Portugal
53942@novasbe.pt

Qiwei Han
*Nova School of Business and Economics*
*Universidade NOVA de Lisboa*
Carcavelos, Portugal
qiwei.han@novasbe.pt

***Abstract*— This study presents a novel application of Graph Neural Networks (GNNs) to optimize dealership network planning for a luxury car manufacturer in the U.S. By conducting a comprehensive literature review on dealership location determinants, the study identifies 65 county-level explanatory variables, augmented by two additional measures of regional interconnectedness derived from social and mobility data. An ablation study involving 34 variable combinations and ten state-of-the-art GNN operators reveals key insights into the predictive power of various variables, particularly highlighting the significance of competition, demographic factors, and mobility patterns in influencing dealership location decisions. The analysis pinpoints seven specific counties as promising targets for network expansion. This research not only illustrates the effectiveness of GNNs in solving complex geospatial decision-making problems but also provides actionable recommendations and valuable methodological insights for industry practitioners.**

## I. Introduction

The automotive industry has witnessed a significant shift toward online vehicle purchasing, prompting manufacturers to explore digital direct-to-consumer (D2C) sales models alongside traditional dealerships. The D2C model enables Original Equipment Manufacturers (OEMs) to capitalize on growing consumer demand for streamlined and cost-effective online sales experiences [1]. However, in two-thirds of U.S. states, franchising regulations mandate that OEMs with existing dealership networks continue to sell vehicles exclusively through these franchised dealerships, which handle all retail activities for customers [2]. Despite the rise of online sales, dealership visits remain important to consumer purchasing decisions, especially in the luxury car segment, where high quality service and repeat business are pivotal revenue drivers [1], [3]. Luxury car dealerships cater to a discerning, often senior clientele by offering personalized services, such as a dedicated sales contact and home pick-up for service appointments [3]. Although the trend toward online sales has reduced the necessity for dense dealership networks, proximity to key markets remains critical [4], [5]. Thus, understanding the factors that influence dealership location decisions and identifying underserved regions is essential for luxury car manufacturers aiming to thrive in the U.S. market.

Location decisions for luxury vehicle dealerships are complex, as they must consider not only the characteristics of a specific geographic area such as demographic factors like population and income at the dealership site [6], [7], [8], [9], but also its connectedness to surrounding regions, such as the site's interconnectedness with neighboring areas, often measured using mobility data [10], [11], [12]. Given the importance of both demographic and interconnectedness data, geographic areas can be represented as nodes in a network, facilitating the prediction of potential new dealership locations.

Recent advances in Graph Neural Networks (GNNs) have significantly improved performance by enabling models to learn from both a node's attributes and its neighborhood. This capability makes GNNs particularly suitable for geospatial inference tasks [14]. In site selection contexts, GNNs have outperformed traditional models that focus solely on local features, though no existing studies have tailored GNN approaches specifically to site selection for luxury retails [15]. Given that retail location decisions are highly contingent on a company's industry and strategy [16], a customized approach to model and feature selection could enhance decision-making processes.

This study introduces a GNN-based approach to identify the drivers of location decisions and predict optimal new dealership sites for luxury car manufacturer Company X in the U.S. Using a county-level dataset, the study builds on a literature review of exogenous factors influencing dealership locations within the luxury automotive sector, which informs the selection of county-level explanatory variables across five categories. Additionally, a thorough review of regional interconnectedness metrics guides the inclusion of mobility and social flow data between counties. An ablation study evaluating the performance of ten state-of-the-art GNN models across 34 variable combinations provides insights into the predictive power of multiple explanatory variables. The best-performing model-variable combination is then applied to counties without any existing Company X dealerships, generating predictions for potential expansion sites.

This research offers valuable insights for discovering valuable places both for Company X and for industry practitioners in a more general sense. The ablation study results and model interpretations deepen the understanding of factors driving the manufacturer's past location choices, particularly the role of demographic and mobility variables. Moreover, the model's predictions, when balanced against the risk of cannibalizing existing sales, provide strategic recommendations for future dealership locations. The methodology can also be adapted and extended to support location decision-making in other retail sectors.

## II. Background and Literature

The existing literature provides a robust foundation for understanding the critical role of local car dealership presence in driving brand success, particularly within geographically defined markets. Extensive research has been conducted to identify the factors that influence dealership performance at specific locations, which is crucial for informed network

To encourage further research, we release our code, models and data at: https://github.com/carocciluca/gnn-site-selection

planning. In this section, we review key studies on the significance of local presence for manufacturer success and examine the variable types commonly employed in evaluating dealership performance and location decisions.

### A. Relevance of Local Presence for Business Success

The presence of local dealerships is intrinsically linked to a brand's success within a specific region. Consumers exhibit a strong preference for purchasing vehicles from dealerships near their homes, with the likelihood of purchase diminishing as the distance between buyer and seller increases [17]. Consequently, automotive brands can enhance their market share by developing concentrated and accessible dealership networks. However, the benefits of increasing network density are subject to diminishing returns, particularly due to the potential for cannibalization of sales among closely located dealerships [18]. A 2021 study of the Texas market revealed that the median distance between a customer and the selling dealership is only 5.2 miles, with the majority of customers purchasing vehicles within 30 miles of their home [19]. Interestingly, over the past five years, the distance consumers are willing to travel to a dealership has increased by 11%, largely due to the widespread availability of online information [4]. Despite this trend, proximity to key markets continues to be a critical success factor for new vehicle dealerships [5]. In addition to proximity, customer satisfaction with dealership services plays a significant role in determining a brand's regional success, as satisfied customers are more likely to return for future purchases and service needs [20].

### B. Location Choices for Automobile Dealerships

Given the critical importance of local presence discussed in the previous section, extensive research has been conducted to analyze the factors influencing dealership location choices. The primary determinants identified in the literature are market size and competition, both of which play significant roles in the performance of car dealerships and, consequently, should be key considerations in expansion strategies. In this section, we review the key explanatory variables used in existing studies on vehicle type choice and the impact of competition on dealership performance. Table 1 categorizes these variables into five main groups: *Basic Demographics*, *Wealth*, *Transportation-Related Consumer Behavior*, *Psychographics*, and *Competition*. The first four groups primarily affect market size, while the fifth focuses on regional competition. Additionally, we examine the importance of regional interconnectedness in location strategy, as summarized in Table 2.

*1) Market Size:* Market size is a foundational determinant in deciding the number and location of car dealerships within a region [6]. Research by Waldfogel highlights how individuals with similar demographic characteristics, such as income, tend to cluster geographically. This clustering leads to more concentrated patterns of expenditure within these groups, which in turn creates demand for products and services tailored to their specific preferences [25]. This finding suggests that retail stores, including car dealerships, are likely to consider the demographic composition of surrounding areas when planning new outlets. Vitorino's study on department stores further supports this idea, showing that population density and the purchasing power of a store's surrounding area positively influence the entry decisions of high-end stores, though they have the opposite effect on non-high-end stores [26].

TABLE 1. TYPES OF EXPLANATORY VARIABLES FROM EXISTING WORKS ON VEHICLE TYPE CHOICES AND THE IMPACT OF COMPETITION ON CAR DEALERSHIPS

| Research Area | Reference | Types of Explanatory Variables | | | | | Dependent Variable |
|---|---|---|---|---|---|---|---|
| | | ***Basic Demographics*** | ***Wealth*** | ***Transportation-Related Consumer Behavior*** | ***Psychographics*** | ***Competition*** | |
| Vehicle Type Choice | [8], [9] | • Gender<br>• Age<br>• Education<br>• Employment<br>• Residential Area<br>• Household Composition | • Income | • Number of Vehicles<br>• Travel Habits (e.g., commuting time & distance, miles traveled by airplane)<br>• Travel-Related Attitudes | • Lifestyle (e.g., Importance of social status) | - | Vehicle Type Choice |
| | [21] | • Gender<br>• Age<br>• Education<br>• Employment<br>• Household Composition | • Income | • Number of Vehicles<br>• Market Price & Age of Household Fleet<br>• Vehicle Types in Household Fleet | - | - | Vehicle Class and Vintage Choice |
| | [7] | • Gender<br>• Education<br>• Residential Area<br>• Marital Status | • Income | • Purpose of Car Use<br>• Driving Skills | • Social Status Involvement | - | Vehicle Type Choice |
| Impact of Competition on Automobile Dealerships | [22] | • Age<br>• Education<br>• Employment<br>• Ethnicity | • Income | • Households Commuting by Public Transportation | - | • Proximity of Competing Dealerships | Inventory (in Days-of-Supply) |
| | [23] | • Education<br>• Residential Area<br>• Household Composition | • Income | - | - | • Proximity of Competing Dealerships | Dealership Profits |
| | [24] | • Employment<br>• Ethnicity | • Income | - | - | • Proximity of Competing Dealerships | Car Retail Prices |

Choo and Mokhtarian analyzed vehicle type choice by examining basic demographic attributes such as gender, age, education, employment, and residential area, alongside income [8]. Their study also included data on travel habits, such as commuting and airplane travel, and lifestyle attitudes, such as the importance of social status. The findings indicate that luxury car owners are more likely to be male, older or retired, highly educated, and high-income earners. Additionally, those who place high importance on perceived social status are more inclined to choose luxury cars. Notably, luxury car owners tend to drive less frequently and opt for air travel more often than other demographic groups [9].

Mohammadian and Miller expanded on these findings by focusing on a similar set of basic demographic variables, in addition to income, and incorporating factors related to a household's existing vehicle fleet, such as the number of vehicles, vehicle types, market price of the fleet, and the age of the cars. While their study did not specifically address luxury vehicle choice, they observed that higher education levels are positively associated with the purchase of new rather than used vehicles [21].

Baltas and Saridakis also used a similar set of demographic and income variables but added information on the purpose of car use, driving skills, and involvement in car-related social status. Their research found that consumers in urban areas show a higher preference for luxury cars. Moreover, they confirmed that social status orientation increases the likelihood of purchasing a luxury car. However, increased car use for commuting was found to reduce the preference for luxury cars [7].

*2) Competition:* The presence of competition in a region is another critical factor in evaluating potential retail locations. A well-documented phenomenon across various sectors, including the automotive dealership industry, is the tendency of similar stores to co-locate with their direct competitors [27]. This behavior, known as agglomeration, has been studied extensively not only in the context of automobile dealerships but also in other retail sectors, such as grocery stores and fast-food restaurants. For instance, Qiaowei and Xiao found that the presence of a competitor positively influences a firm's decision to expand within the fast-food industry [28]. Similarly, Chidambaram and Pervin observed that restaurants operating in areas with competitors in a similar price range experience significantly lower failure rates [29].

The potential benefits of co-location for car dealerships are manifold. First, firms can achieve economies of advertising and promotion by gaining exposure to potential customers who visit competing firms in the same area. Second, clustering multiple franchises of the same firm can lead to increased operational efficiencies and scale economies, such as joint ordering from manufacturers or consolidated brand advertising efforts. Third, both firms and consumers benefit from reduced uncertainty. Firms can learn and adapt to new markets by observing their competitors' entry behavior, while consumers can overcome taste uncertainty at lower search costs by physically comparing products at nearby outlets [27]. Agglomeration forces are particularly strong for products that require customers to visit multiple stores to try and compare options [30].

Despite the advantages of agglomeration, studies assessing its impact on business performance in the automobile dealership sector have yielded mixed results. Olivares and Cachon indicated that competition in isolated U.S. markets could negatively affect dealership performance in two ways. First, increased competition may reduce individual dealership sales. Second, dealerships may respond by enhancing service levels, such as expanding car availability, to attract new customers, which could lead to an increase in inventory relative to sales [22]. Conversely, Murry and Zhou found that while increased competition drives dealerships to lower prices, the benefits of agglomeration, such as higher sales volumes driven by increased consumer traffic, can outweigh the disadvantages [23]. Beard, et al., provided evidence that dealerships charge lower prices when other dealerships sell the same model, indicating strong intra-brand competition. Interestingly, however, they did not find this effect in the presence of inter-brand competition [24]. Although none of these studies specifically focused on the luxury car segment, it is likely that economies of agglomeration play a crucial role in expansion decisions within this segment as well.

*3) Regional Connectedness:* Regional connectedness is a critical factor in retail location strategy, particularly for sectors dealing with high-value goods like luxury vehicles. The degree of interconnectedness between geographic regions provides insights into potential dealership locations that could serve not just the immediate area but also neighboring regions. Verhetsel et al., utilized consumer flow data to delineate communities and identify hubs that supply various goods and services to their respective communities. Their findings indicate that the size and influence of a community vary depending on the type of goods being analyzed; for example, car purchases tend to create larger community hubs compared to daily goods like groceries [10]. This highlights the importance of considering not only the demographic and competitive factors of a potential dealership site but also its connectivity to surrounding areas.

As shown in Table 2 academia, government agencies, and companies have developed and utilized a variety of indicators to measure regional connectedness. These indicators include metrics such as consumer retail flows, commuting flows, and social connectedness indices, which are becoming increasingly important in the era of big data [31]. These metrics are vital for identifying regions that are not only demographically promising but also strategically positioned within broader mobility and social networks, enhancing the potential success of new dealership locations.

TABLE 2. INDICATORS OF REGIONAL CONNECTEDNESS

| Indicator | Explanation | Source | Reference |
|---|---|---|---|
| Consumer Retail Flows | # of people from region *i* shopping for given goods in region *j* | Survey | [10] |
| Commuting Flows | # of people regularly commuting from region *i* to region *j* for work | Survey[a] | [11] |
| Population Flows | # of people from region *i* that visited region *j* within a given time interval | Mobile GPS | [12] |
| Social Connected-ness Index (SCI) | The relative probability that a Facebook user in region *i* has a friendship link to any user in region *j* | Social Media[a] | [31] |

[a]. Data is publicly available in full.

## C. Graph Neural Networks and Geospatial Applications

Graph structures provide a powerful framework for representing data across a wide range of domains, particularly when interactions or relationships between entities need to be modeled. The fundamental components of a graph are nodes and edges: nodes represent entities, and edges represent the relationships or interactions between these entities. Both nodes and edges can have attributes associated with them. For example, in a social media context, users are often represented as nodes, with edges representing friendships between users. Node attributes might include details such as age, gender, and interests, while edge attributes might capture the nature and frequency of interactions, such as likes or messages exchanged.

Traditionally, processing data represented as graphs posed significant challenges because it required converting graph-based data into simpler structures, like vectors, to be compatible with conventional machine learning algorithms [13]. This conversion often led to the loss of crucial structural information, such as the relationships and dependencies inherent in the graph. To address this issue, Scarselli et al. introduced the Graph Neural Network (GNN), a model capable of directly processing graph structures for both graph-focused and node-focused tasks [32]. Initially, GNNs were built on Recurrent Neural Network (RNN) architectures, but subsequent developments have seen the adoption of Convolutional Neural Network (CNN) architectures, which have significantly improved the speed and accuracy of GNNs [33].

The primary advantage of GNNs lies in their ability to learn not only from the attributes of individual nodes but also from the attributes and structure of their neighboring nodes. This capability makes GNNs particularly well-suited for geospatial inference tasks, where the relationship between a location and its surrounding areas is critical. For example, Zhu et al. utilized Graph Convolutional Networks (GCNs), a specific GNN type, to predict unknown functional characteristics of locations by leveraging both the known characteristics of these locations and their connections to other locations [14]. Other notable applications of GNNs in geographic contexts include point-of-interest recommendations [34] and unsupervised approaches for community detection within spatial networks [35].

In the context of site selection, GNNs offer significant advantages over traditional methods, which often struggle with the complexity of incorporating spatial relationships or tend to focus narrowly on local features without considering broader regional interactions. For instance, Lan et al. proposed a GCN-based approach to assess the attractiveness of areas using Google Maps store ratings, utilizing a comprehensive dataset that included traffic, business, and residential data [15]. However, their approach was general and did not account for industry-specific factors, nor did it evaluate different model or feature combinations tailored to specific retail types. As site selection strategies are highly dependent on the specific industry and company strategy, a tailored approach to both model and feature selection is essential for making informed decisions [16].

Furthermore, many existing studies do not incorporate critical demographic data, which plays a significant role in site selection decisions [26]. By representing geographic locations as nodes and incorporating both node attributes (e.g., demographic characteristics) and edge attributes (e.g., mobility flows or social connections), GNNs provide a more holistic approach to site selection. This approach allows for the integration of a wide range of data sources, enhancing the predictive power and relevance of the models used for determining optimal dealership locations.

# III. Methodology

## A. Notations and Problem Definition

Formally, we define a graph $G = (V, E)$ where $V$ represents the set of nodes $V = (v_1, \dots, v_n)$ , each node corresponding to one U.S. county. The set $E$ denotes the edges $E = \{e_{ij} = (v_i, v_j)\}$, representing the connections between pairs of counties. Here, $n = |V|$ is the total number of nodes (counties), and $m = |E|$ is the total number of edges in the dataset. The node feature matrix $\boldsymbol{X} \in \mathbb{R}^{n \times d}$ captures the $d$ demographic attributes associated with each county. The node label matrix $\boldsymbol{L} \in \mathbb{R}^{n \times 1}$ contains binary ground-truth labels $l_i \in \{0,1\}$ where $l_i = 1$ indicates that county $v_i$ has at least one Company X dealership, while $l_i = 0$ signifies the absence of a dealership. The edge feature matrix $\boldsymbol{E} \in \mathbb{R}^{m \times k}$ encodes the $k$ features associated with each connection between counties.

The objective of the binary classification model is to predict whether a specific U.S. county $(v_i)$, for which the presence of a Company X dealership $(l_i)$ is unknown, either hosts a dealership ($l_i = 1$) or does not ($l_i = 0$). To identify potential targets for dealership expansion, we reserve a random sample of counties with $l_i = 0$ during the training, validation, and testing phases. After training, the model is applied to this sample to predict the likelihood of a dealership's presence. If the model predicts $\hat{l}_i = 1$ for a county, that county $(v_i)$ is flagged as a potential expansion target.

This model bases its recommendations on historical site selection patterns of the manufacturer. Given the absence of sensitive financial data, the model operates under the assumption that replicating past decisions is beneficial. However, to enhance decision-making, we suggest reformulating the problem as a regression task that predicts a performance metric, such as dealership profitability. This approach would allow the model not only to learn from historical decisions but also to evaluate the potential profitability of expanding into counties without an existing dealership.

## B. Dataset Description and Preprocessing

*1) County Variables:* County-level data was primarily sourced from SimplyAnalytics, a data analytics and visualization platform that aggregates demographic, business, and marketing information from multiple data partners [36]. A total of 65 county-level explanatory variables were selected, reflecting the categories identified in the literature. These variables are detailed in Table 3, along with their respective descriptions..

- **Basic Demographics (BD):** This category includes variables that describe the fundamental demographic characteristics of a county. Examples include population density, household size distribution, the proportion of family households, age distribution, and the percentage of the population holding a college degree or higher. Variables such as household size and age distribution are broken down into specific ranges, such as the percentage of the population aged 25 to 34 years.

- **Wealth (WE):** Represents different wealth measures such as income distribution, net worth distribution, unemployment rate, and poverty rate. In addition, the category includes variables on renter-occupied housing and housing mortgages as these two measures may influence discretionary income significantly.
- **Transportation Behavior (TB):** Comprises variables measuring transportation-related consumer behavior such as the number of vehicles owned, commuting behavior, as well as household spending on new vehicles and airline fares.
- **Luxury Behavior (LB):** Derived from the "Psychographics" category identified in the literature review. It contains variables that measure relevant psychographic attributes (e.g., share of "auto luxury lovers") assessed by MRI-Simmons. Other variables measure luxury-related consumption behavior such as the population share that visited fine dining restaurants over the past month. The local Google Trends Index for the keyword "Company X" completes this category [37]. Since the index is only provided on a metropolitan area level, we mapped metro-level values to a county level.
- **Competition (CO):** Contains two binary variables that denote whether a county has a Competitor A or Competitor B dealership. We consider these two brands Company X's main competitors due to their luxury positioning, country origin, and comparable retail footprint in the U.S. market. We collected dealership locations from the company websites and retrieved coordinates via the Google Maps API [38].

The dataset for analysis comprised 3,105 U.S. counties, excluding non-contiguous states and outlying areas. Further, we excluded the three counties with a population below 300 as they appeared as strong outliers across multiple variables, potentially indicating poor data quality. We interpolated missing values using averages of adjacent counties, drawing from the First Law of Geography [39]. Lastly, we addressed the general data-related issues of skewness and data heterogeneity [40]. Variables with skewed distributions were transformed using the Yeo-Johnson technique and all variables were scaled to values between 0 and 1.

*2) County-to-County Connection Variables:* The connectivity between counties plays a crucial role in dealership network planning, particularly when a single dealership can potentially serve multiple interconnected counties. To capture this, we selected two key variables representing social and mobility connectedness between counties.

- **Social Connectedness:** The Social Connectedness Index (SCI), provided by Meta, measures the strength of social ties between two regions based on the likelihood that a Facebook user in one region is friends with a user in another region [41].
- **Mobility Connectedness** This metric assesses mobility between counties using data on county-to-county commuting flows from the American Community Survey [11]. We define the Mobility Connectedness Index (MCI) between counties *i* and *j* as

$$MCI_{ij} = \left( \frac{\frac{Commuting\ Workers_{ij}}{Population_i}}{max_{k,l}\left(\frac{Commuting\ Workers_{kl}}{Population_k}\right)} \right) \quad (1)$$

where *(k, l)* represents all county pairs in the dataset.

### C. Modeling and Evaluation

To evaluate the effectiveness of different county and county-to-county connectedness variables, we conducted an ablation study using ten state-of-the-art GNN operators from the PyTorch Geometric library [42]. Each model was initialized with a single convolutional layer, mapping the *d* demographic node features to a one-dimensional output. This was followed by a sigmoid activation function to convert the logits into probabilities ranging from 0 to 1. The model predicts the presence of a Company X dealership in a county if the output value for the corresponding node is ≥0.5. For GNNs that required a neural network as an input parameter, we implemented a single linear layer as the neural network. Table 3 provides a comprehensive overview of all the algorithms and parameters utilized in this study.

TABLE 3. GNN ALGORITHMS USED IN ABLATION STUDY

| GNN Algorithm | Specified Parameters[a] | Source |
|---|---|---|
| Residual Gated Graph Convolutional Network (ResGatedGraphConv) | • in_channels: # of node features<br>• out_channels: 1<br>• edge_dim: # of edge features | [33] |
| Graph Attention Networks (GATConv) | • in_channels: # of node features<br>• out_channels: 1<br>• edge_dim: # of edge features | [43] |
| Graph Attention Networks v2 (GATv2Conv) | • in_channels: # of node features<br>• out_channels: 1<br>• edge_dim: # of edge features | [44] |
| Graph Transformer (TransformerConv) | • in_channels: # of node features<br>• out_channels: 1<br>• edge_dim: # of edge features | [45] |
| Graph Isomorphism Network (GINEConv) | • nn: a Linear layer with *in_features = # of node features* and *out_features = 1*<br>• edge_dim: # of edge features | [46] |
| Gaussian Mixture Model Convolutional Network (GMMConv) | • in_channels: # of node features<br>• out_channels: 1<br>• dim: # of edge features<br>• kernel_size: 5 | [47] |
| Message Passing Neural Networks (MPNN) | • in_channels: # of node features<br>• out_channels: 1<br>• nn: a Linear layer with *in_features = # of edge features* and *out_features = # of node features* | [48] |
| GENeralized Graph Convolution Network (GENConv) | • in_channels: # of node features<br>• out_channels: 1<br>• edge_dim: # of edge features | [49] |
| Pathfinder Discovery Network (PDNConv) | • in_channels: # of node features<br>• out_channels: 1<br>• edge_dim: # of edge features<br>• hidden_channels: # of edge features | [50] |
| General GNN (GeneralConv) | • in_channels: # of node features<br>• out_channels: 1<br>• in_edge_channels: # of edge features | [51] |

[a] All optional, unmentioned parameters are set to their standard configurations.

Prior to training, we set aside 20% of counties without a Company X dealership (563 counties in total) as an "application set." The purpose of this application set was to deploy the best-performing model-feature combination to predict potential expansion targets. The remaining dataset was split into five equal folds, and stratified k-fold cross-validation was employed to ensure consistent representation of counties with and without a Company X dealership across all folds. During each iteration of the cross-validation, the model was trained on three folds for up to 2,000 epochs. To mitigate overfitting, we monitored the binary cross-entropy loss on the validation fold during each epoch and applied early stopping if no improvement was observed for 50 consecutive epochs. After training, the model was tested on the test fold for that iteration, and performance metrics were recorded. The process was repeated for all five folds, and the performance metrics were averaged to compare different model-feature combinations

We selected the F-beta score with parameter $\beta = 1/3$ as the primary performance metric to avoid the limitations of accuracy, particularly in imbalanced datasets. The F-1/3 score score places greater emphasis on precision, which is critical given the high cost of investigating potential expansion targets. Nevertheless, recall is also considered, penalizing models that overly favor the majority class. A higher F-1/3 score indicates a model that, while conservative, makes more reliable predictions, leading to fewer but more focused recommendations for expansion.

The ablation study was conducted in two rounds. In the first round, we evaluated the predictive power of each variable group by running the k-fold cross-validation process on each group in combination with each of the ten GNN algorithms. The variable groups were then ranked by computing the median F-1/3 scores across all algorithms. As a baseline, we defined edges based on geographic adjacency (i.e., counties are connected if they are geographically adjacent) and included both the SCI and MCI as edge features. Finally, we selected the best model-feature combination and deployed it on the application set to predict potential expansion targets. We also applied a Captum-based explainer to interpret the model's predictions.

In the second round, we reassessed the performance of all ten algorithms by varying the node feature combinations, gradually adding feature groups in descending order of their median performance from the first round. We also experimented with different edge feature combinations by assessing SCI and MCI individually and in combination, using two edge definitions: one based solely on geographic adjacency and another based on all counties with non-zero commuting flows. The underlying hypothesis for this latter definition was that consumers might be willing to travel to a dealership in another county if it falls within their commuting range.

The best-performing model-feature combination from the ablation study was deployed on the application set to predict potential dealership expansion targets. To interpret the model's predictions, we used Captum, a model interpretability tool, to analyze the contributions of different features to the model's decisions.

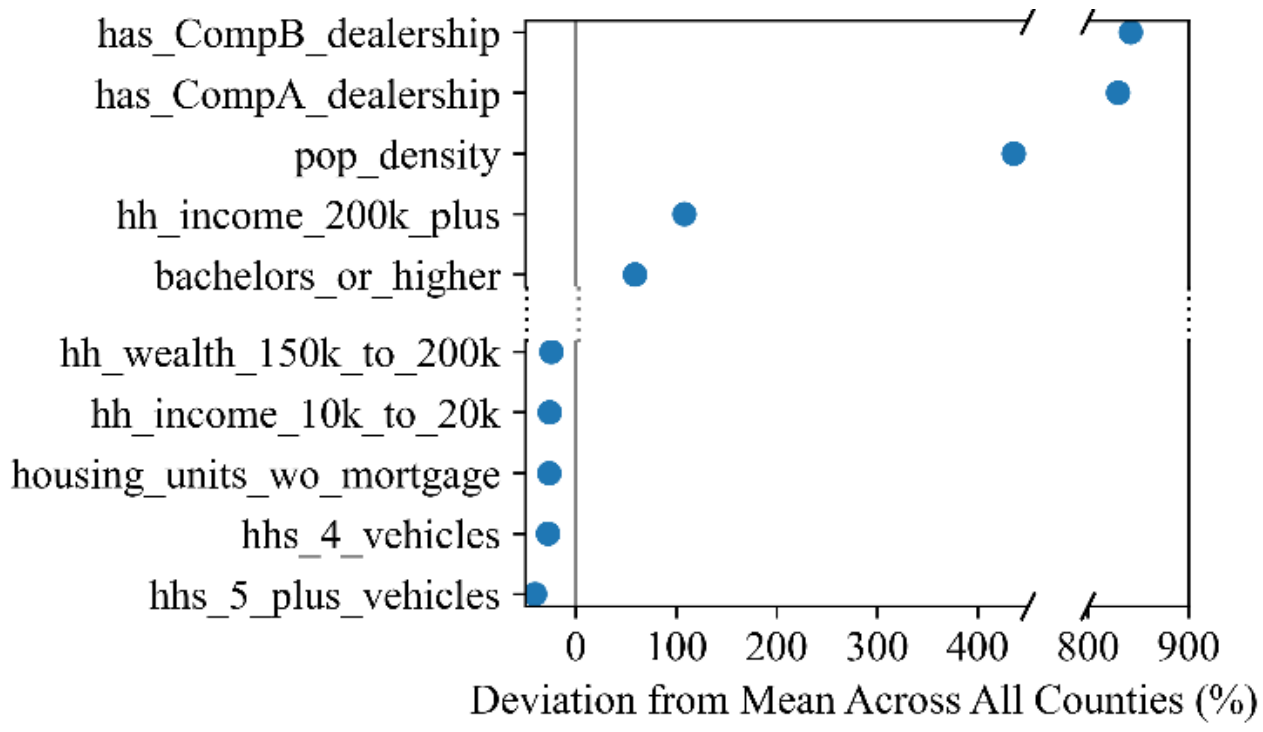

Fig. 1. Top five most positive (negative) deviations of variable averages for counties with Company X presence compared to all counties

## IV. Exploratory Data Analysis

The final dataset comprises 3,105 U.S. counties, each described by 66 variables, including the presence of Company X dealerships. At the time of analysis, Company X had dealerships in only 9.3% of counties, highlighting a significant class imbalance that could impact model performance. Figure 1 illustrates the differences between counties with a Company X dealership and the overall dataset averages. This comparison reveals that counties hosting a Company X dealership are significantly more likely to also have a Competitor A/B dealership—over eight times more likely than the average county. This strong co-location pattern suggests that competition variables may be more predictive of dealership presence than traditional wealth indicators, such as the proportion of households earning over $200,000 annually, which is only twice as high in counties with a Company X dealership compared to the average.

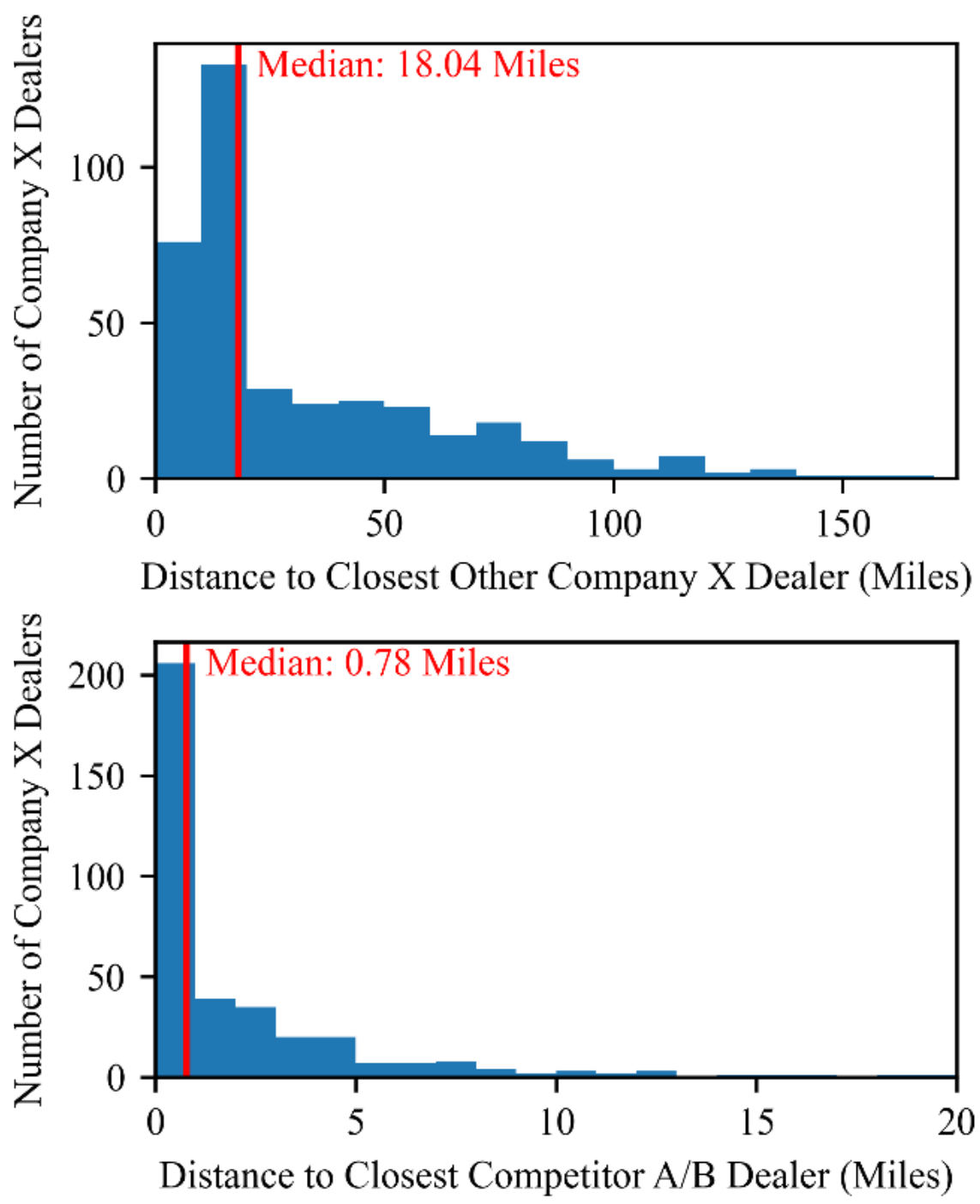

Fig. 2. Distributions of geodesic distances between dealerships

TABLE 4. TOP 10 MODEL-FEATURE COMBINATIONS OF ABLATION STUDY BY AVERAGE F-1/3 SCORE

| Model | Edge Definition | Edge Features | Node Features[a] | F-1/3 | Precision | Recall | Accuracy |
|---|---|---|---|---|---|---|---|
| MPNN | Adjacent | MCI | CO + BD + WE + LB | 0.92 | 0.94 | 0.74 | 0.97 |
| GeneralConv | Adjacent | SCI + MCI | CO + BD + WE + LB | 0.91 | 0.93 | 0.74 | 0.96 |
| GeneralConv | Adjacent | MCI | CO + BD + WE | 0.91 | 0.93 | 0.75 | 0.96 |
| GeneralConv | Adjacent | MCI | CO + BD + WE + LB + TB | 0.90 | 0.92 | 0.77 | 0.97 |
| ResGatedGraphConv | Adjacent | MCI | CO + BD + WE | 0.90 | 0.91 | 0.83 | 0.97 |
| GMMConv | Commuting Flows | SCI + MCI | CO + BD + WE | 0.90 | 0.90 | 0.84 | 0.97 |
| MPNN | Adjacent | SCI + MCI | CO + BD + WE | 0.90 | 0.91 | 0.81 | 0.97 |
| TransformerConv | Adjacent | MCI | CO + BD + WE + LB | 0.90 | 0.91 | 0.82 | 0.97 |
| GMMConv | Commuting Flows | MCI | CO + BD + WE + LB + TB | 0.90 | 0.91 | 0.80 | 0.97 |
| ResGatedGraphConv | Commuting Flows | MCI | CO + BD + WE + LB + TB | 0.89 | 0.90 | 0.83 | 0.97 |

[a] Legend: BD = Basic Demographics, WE = Wealth, TB = Transportation Behavior, LB = Luxury Behavior, CO = Competition.

Figure 2 presents the density of the existing Company X dealership network. The median geodesic distance between dealerships is 18 miles, a metric that should be carefully considered in any expansion strategy to avoid oversaturating the market. The analysis also reveals a strong co-location tendency, with the median Company X dealership situated within walking distance of a competitor's dealership. This proximity showcases the importance of competition variables in predicting dealership locations and highlights the strategic importance of co-location in the luxury automotive market.

## V. RESULTS AND INTERPRETATIONS

In the first round of the ablation study, we evaluated the performance of all ten models using each variable group independently. The results revealed that Competition (CO) variables were the most predictive of Company X dealership locations, achieving a median F-1/3 score of 0.88 across all models. Basic Demographics (BD) and Wealth (WE) followed with F-1/3 scores of 0.70 and 0.69, respectively. Based on these findings, we conducted the second round of the ablation study by gradually combining the variable groups, starting with Competition variables as the baseline and sequentially adding Basic Demographics, Wealth, Luxury Behavior (LB), and Transportation Behavior (TB) variables.

Although Competition variables alone showed strong predictive power individually, the top ten model-feature combinations in the second round did not rely solely on them. As shown in Table 4, all top-performing combinations required at least a mix of Competition, Basic Demographics, and Wealth variables to achieve optimal results. This suggests that while co-location effects are significant for luxury car dealerships, manufacturers likely consider a broader range of factors when planning their networks, factors that may vary across different brands.

The analysis of edge features revealed key insights into consumer behavior and its impact on dealership location decisions. First, the Mobility Connectedness Index (MCI) was included in all top 10 model configurations, with seven models relying exclusively on MCI. This finding emphasizes the importance of physical mobility between counties in dealership network planning. Second, seven out of the top ten models used geographic adjacency as a basis for defining edges, further reinforcing the idea that physical proximity between counties plays a more significant role than social proximity in location decisions. Among all models tested, MPNN using MCI-based adjacency edges, and a combination of Competition, Basic Demographics, Wealth, and Luxury Behavior variables achieved the highest F-1/3 score. While MPNN ranked first in precision (0.92), it placed ninth in recall (0.74), indicating that the model tends to make conservative predictions—only predicting dealership presence when it is highly certain. As shown in Table 5, MPNN's performance benefited significantly from the use of adjacency-based edges, outperforming the baseline model using only Competition features. Applying MPNN to the application set, which consisted of counties without a Company X dealership, the model predicted dealership presence (nodes highlighted with black boarder) in nine counties—approximately 1.6% of the 563 counties. Figure 3 visualizes the output node embeddings generated by the model, showing that the model effectively clustered counties based on their likelihood of hosting a dealership.

TABLE 5. AVERAGE F-1/3 SCORES ACHIEVED BY NNCONV FOR ALL TESTED FEATURE COMBINATIONS

| Edge Definition | Edge Features | Node Feature Combinations[b] | | | | |
|---|---|---|---|---|---|---|
| | | *CO* | *CO + BD* | *CO + BD + WE* | *CO + BD + WE + LB* | *CO + BD + WE + LB + TB* |
| **Adjacent** | ***MCI*** | 0.880 | 0.874 | 0.876 | **0.917** | 0.832 |
| | ***SCI*** | 0.879 | 0.877 | **0.880** | 0.873 | 0.860 |
| | ***SCI + MCI*** | 0.880 | 0.881 | **0.896** | 0.853 | 0.811 |
| **Commuting Flows** | ***MCI*** | **0.883** | 0.752 | 0.795 | 0.705 | 0.743 |
| | ***SCI*** | **0.882** | 0.760 | 0.771 | 0.730 | 0.684 |
| | ***SCI + MCI*** | **0.883** | 0.787 | 0.755 | 0.702 | 0.664 |

[a] Node (Edge) feature group that led to best performance for each edge (node) feature group in bold (underlined).
[b] Legend: BD = Basic Demographics, WE = Wealth, TB = Transportation Behavior, LB = Luxury Behavior, CO = Competition.

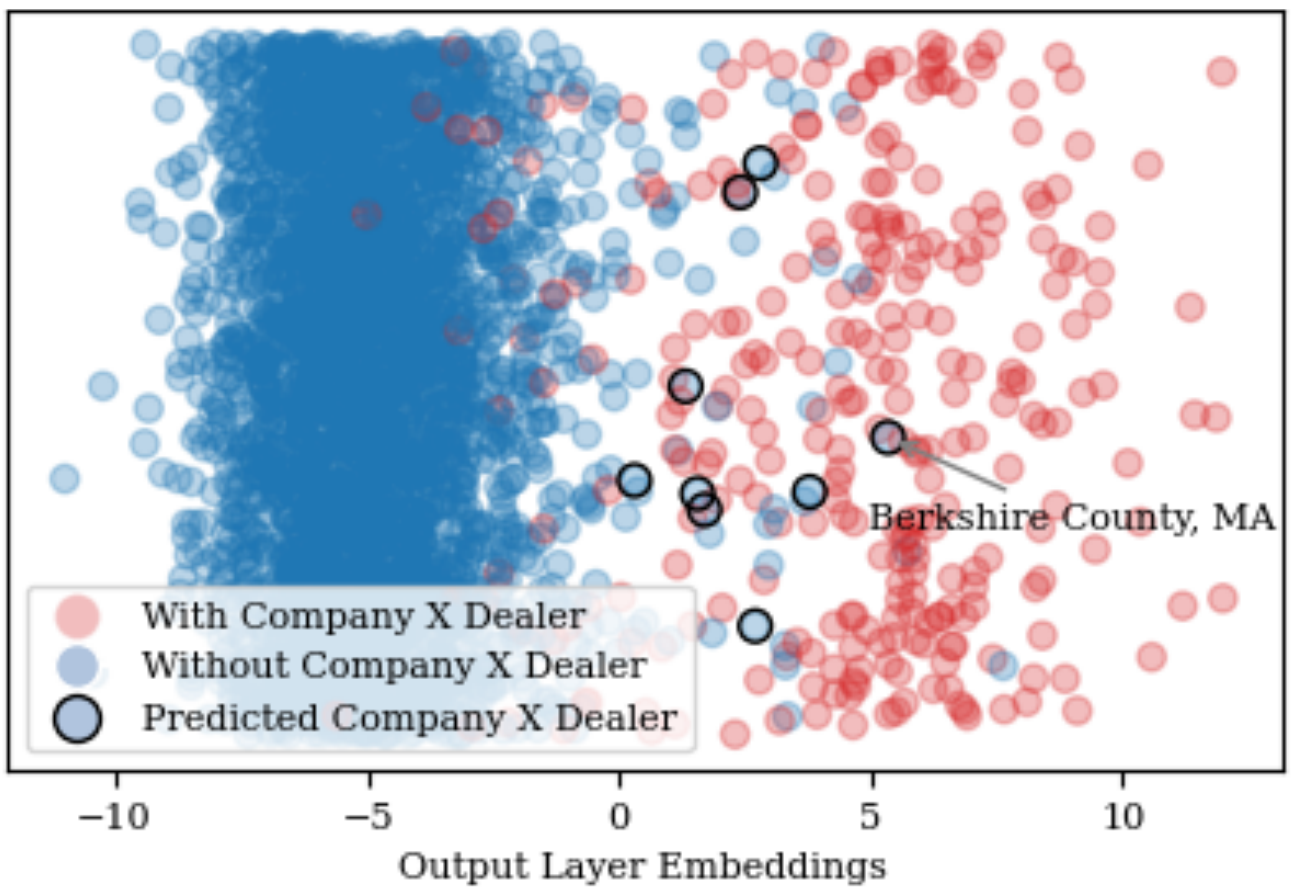

Fig. 3. Output node embeddings generated by MPNN with best feature combination (Note: output layer embeddings are one-dimensional and randomly jittered across y-axis for visualization purposes only)

Table 6 lists the counties where the model predicted a dealership presence, serving as potential expansion targets. Notably, eight out of the nine counties already have at least one competitor dealership, underscoring the strong co-location tendency. The ninth prediction, Houston County (GA), had no competitor presence but was still flagged by the model, suggesting other influential factors. Captum-based feature importance analysis confirmed that competitor presence and population density were consistently among the top predictors for these counties, aligning with earlier findings from the exploratory data analysis and the first round of the ablation study.

Fig. 4 shows results from feature importance analysis for two selected counties, and it also highlighted the significant role of income and net worth variables in the model's predictions. For instance, the prediction for Houston County (GA) heavily relied on the share of households with a net worth below $50,000. This reliance on lower net worth variables suggests that the model factors in a broader range of economic conditions beyond just high-income indicators. Despite this, income-related variables, particularly the share of households earning between $150,000 and $200,000, were still among the top ten most influential features for most predictions. Within the luxury behavior category, interest in Company X, as measured by Google search trends, and the share of the population identified as "auto-luxury-lovers" consistently influenced predictions.

To evaluate the risk of cannibalizing existing sales, we assessed the distance from the identified potential locations to the nearest existing Company X dealership. As shown in Table 6, seven out of the nine predicted locations maintain a distance greater than the current median of 18 miles between existing dealerships, suggesting that these locations would not significantly increase network density. Based on these findings, the following seven counties are recommended for further investigation as potential expansion targets (in order of model certainty): *Berkshire County (MA), Pitt County (NC), Marion County (FL), Tulare County (CA), Lauderdale County (MS), Broome County (NY), and Houston County (GA).* Before making final decisions, Company X should conduct additional market analyses and feasibility studies addressing legal, operational, and financial considerations.

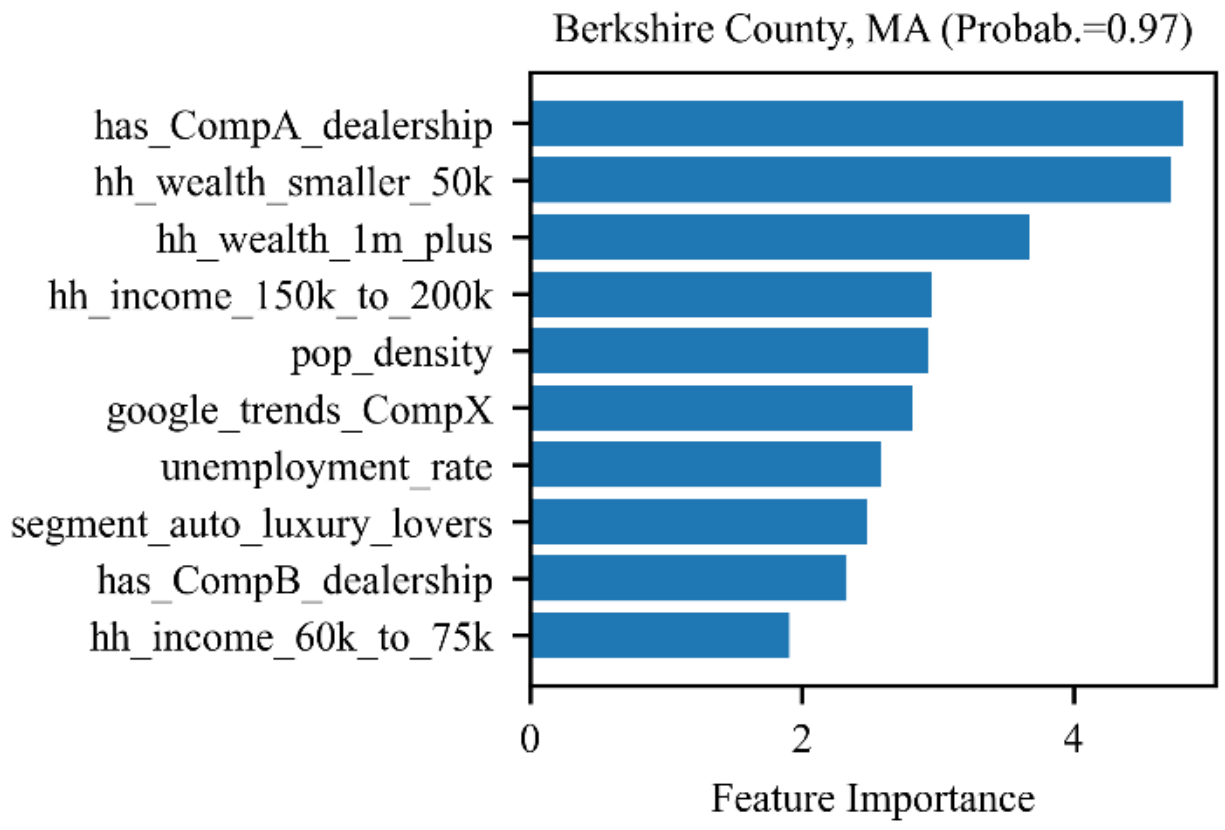

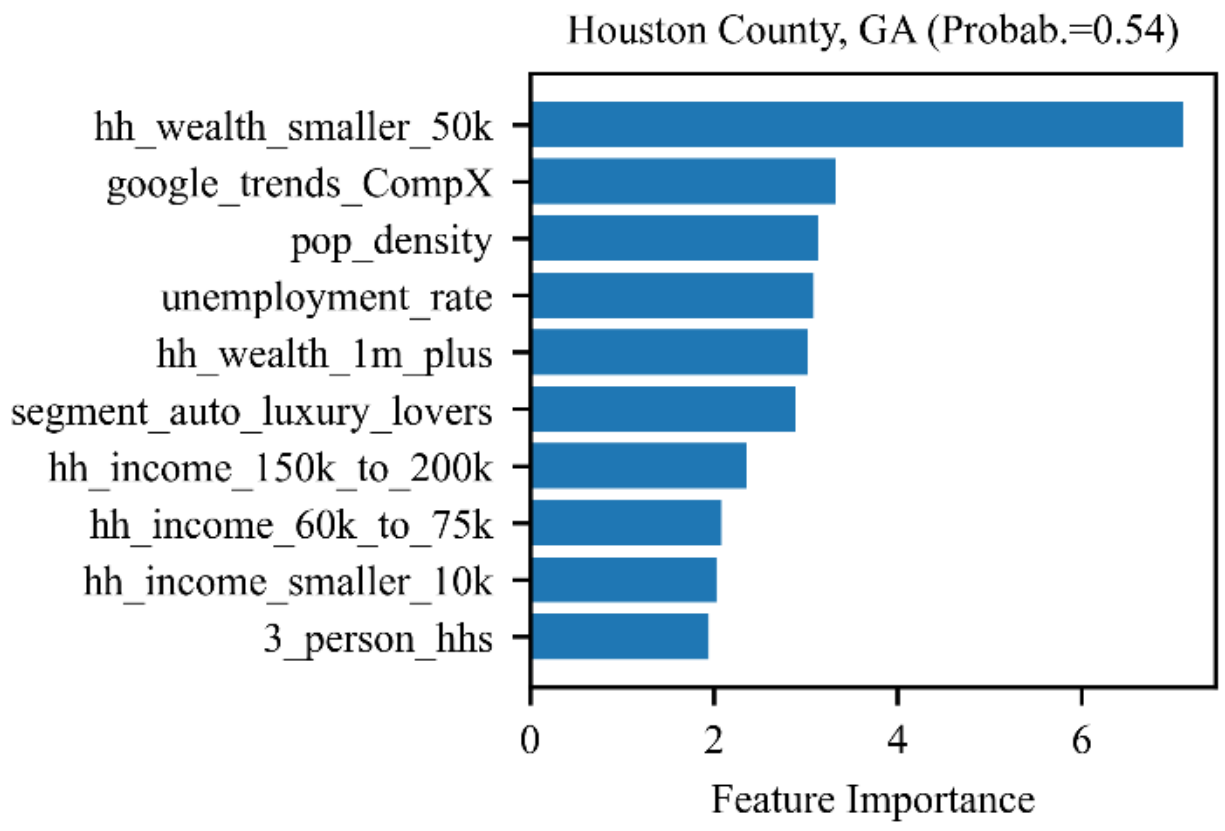

Fig. 4. Feature importances computed by Captum-based explainer for two selected counties

TABLE 6. COUNTIES FROM APPLICATION SET WHERE NNCONV PREDICTED A COMPANY X DEALERSHIP

| FIPS Code | County Name | Predicted Probability of Company X Presence | Current Competitor Presence | Closest Current Company X Dealer (Miles)[a] | Increases Current Network Density[b] |
|---|---|---|---|---|---|
| 25003 | Berkshire County, MA | 0.97 | Competitor A & Competitor B | 37.4 | No |
| 08013 | Boulder County, CO | 0.92 | Competitor A | 13.5 | Yes |
| 37147 | Pitt County, NC | 0.81 | Competitor B | 70.4 | No |
| 12083 | Marion County, FL | 0.76 | Competitor A | 39.1 | No |
| 06107 | Tulare County, CA | 0.69 | Competitor A | 42.1 | No |
| 28075 | Lauderdale County, MS | 0.62 | Competitor A | 84.4 | No |
| 13059 | Clarke County, GA | 0.61 | Competitor A | 2.7 | Yes |
| 36007 | Broome County, NY | 0.55 | Competitor A | 59.7 | No |
| 13153 | Houston County, GA | 0.54 | - | 21.6 | No |

a. Distances calculated assuming full co-location with competitors.

b. Expansion locations identified by model are considered to increase network density if they are located closer to an operating dealer than the current median distance between Company X dealers (18 miles).

## VI. Limitations and Areas for Further Research

While this study demonstrates the potential of Graph Neural Networks (GNNs) in the context of dealership location identification, it is important to acknowledge its limitations, which present avenues for further research. The GNN-based approach, while tailored to the luxury automotive sector in this study, could be adapted and extended to other retail segments. Additionally, there are three key limitations in this work that provide opportunities for future research and refinement.

First, the study's reliance on publicly available mobility data restricted the analysis to using county-to-county commuting flows and the Facebook Social Connectedness Index (SCI). Although mobility data provided stronger model performance compared to social connectedness data, the scope of mobility data was limited. Future research could enhance model accuracy by incorporating more detailed mobility datasets, such as consumer retail flows or population movement data, as has been effectively employed in previous studies leveraging mobility data [10], [12]. Access to such data could provide a more granular understanding of consumer behavior and improve the precision of dealership location predictions.

Second, the lack of access to sensitive financial information necessitated framing the problem as a binary classification task, where the presence of Company X dealerships served as the dependent variable. This approach inherently bases predictions on the manufacturer's historical site selection practices without evaluating the desirability of replicating these decisions. To increase the model's practical relevance, future studies should consider reformulating the problem as a regression task that predicts direct performance indicators, such as dealership profitability or overall revenue contribution. Such a model could better inform location decisions by incorporating financial outcomes and adjusting for endogenous factors, including local employment levels and potential marketing investments by the manufacturer.

Third, while counties provide a relatively granular segmentation of the U.S. territory, they are not uniform in geographic size. This discrepancy could lead to the model overlooking potentially attractive markets within large, predominantly rural counties. Analyzing data at the zip code level or using a custom grid of equally sized geographic areas would offer a finer granularity but presents challenges due to limited data availability at such levels. Future research could explore the impact of advanced interpolation techniques for feature generation on more granular geographic levels, potentially improving the model's ability to identify optimal dealership locations.

## VII. Conclusion

This study has investigated the potential of Graph Neural Networks (GNNs) to identify new locations for expanding the U.S. dealership network of luxury car manufacturer Company X, using county-level data as the basis for analysis. A thorough literature review established that market size—determined by factors such as demographics, wealth, transportation behavior, and luxury consumption patterns—as well as competitor presence, are key determinants influencing the location and performance of luxury automobile dealerships. Moreover, the ablation study demonstrated that competitor presence is a highly effective predictor of Company X's dealership locations, confirming the strong co-location tendencies observed in the luxury automotive sector. Meanwhile, the inclusion of other variable groups, particularly demographics, wealth, and luxury behavior, further enhanced the model's performance, suggesting that Company X's historical location decisions also involve unique considerations beyond those of its competitors. A Captum-based interpretation of the final model's predictions highlighted that Competitor A's presence was the most influential factor, alongside variables related to net worth, unemployment rates, and population density. Lastly, the study emphasizes the importance of the data utilized to represent regional connectedness. Specifically, mobility data—particularly commuting flows—led to significantly better model performance than social connectedness data.

The final model identified seven counties as potential expansion targets that would not exacerbate the existing dealership network density, as these locations would maintain or exceed the current median distance between Company X dealerships. This outcome suggests that the GNN-based methodology employed in this study provides actionable insights that can inform the manufacturer's strategic decisions regarding network expansion. In conclusion, this research demonstrates the utility of GNNs in addressing the complexities of site selection for specific retailers. The proposed methodology not only supports data-driven decision-making for Company X but also presents a scalable framework that could be adapted for use by other companies and sectors. Future research can build on this foundation by exploring the identified limitations and extending the approach to different contexts.

## Acknowledgements

This work was funded by Fundação para a Ciência e a Tecnologia (UIDB/00124/2020, UIDP/00124/2020 and Social Sciences DataLab - PINFRA/22209/2016), POR Lisboa and POR Norte (Social Sciences DataLab, PINFRA/22209/2016).